\documentclass[sigconf]{acmart}
\acmSubmissionID{277}

\AtBeginDocument{%
  \providecommand\BibTeX{{%
    \normalfont B\kern-0.5em{\scshape i\kern-0.25em b}\kern-0.8em\TeX}}}

\copyrightyear{2022}
\acmYear{2022}
\setcopyright{acmcopyright}\acmConference[SA '22 Conference Papers]{SIGGRAPH
Asia 2022 Conference Papers}{December 6--9, 2022}{Daegu, Republic of Korea}
\acmBooktitle{SIGGRAPH Asia 2022 Conference Papers (SA '22 Conference Papers),
December 6--9, 2022, Daegu, Republic of Korea}
\acmPrice{15.00}
\acmDOI{10.1145/3550469.3555393}
\acmISBN{978-1-4503-9470-3/22/12}

\begin{document}

\title{Masked Lip-Sync Prediction by Audio-Visual Contextual Exploitation in Transformers}

\author{Yasheng Sun}
\authornote{Equal contribution.}
\affiliation{%
  \institution{Tokyo Institute of Technology}
  \city{Tokyo}
  \country{Japan}}
\email{sun.y.aj@m.titech.ac.jp}

\author{Hang Zhou}
\authornotemark[1]
\affiliation{%
  \institution{Baidu Inc.}
  \city{Shanghai}
  \country{China}}
\email{zhouhang09@baidu.com}
\author{Kaisiyuan Wang}
\affiliation{%
  \institution{The University of Sydney}
  \city{Sydney}
  \country{Australia}}
\email{kaisiyuan.wang@sydney.edu.au}

\author{Qianyi Wu}
\affiliation{%
  \institution{Monash University}
  \city{Melbourne}
  \country{Australia}}
\email{qianyi.wu@monash.edu}
\author{Zhibin Hong}
\affiliation{%
  \institution{Baidu Inc.}
  \city{Shenzhen}
  \country{China}}
\email{Zhib.hong@gmail.com}

\author{Jingtuo Liu}
\affiliation{%
  \institution{Baidu Inc.}
  \city{Beijing}
  \country{China}}
\email{liujingtuo@baidu.com}

\author{Errui Ding}
\affiliation{%
  \institution{Baidu Inc.}
  \city{Beijing}
  \country{China}}
\email{dingerrui@baidu.com}

\author{Jingdong Wang}
\affiliation{%
  \institution{Baidu Inc.}
  \city{Beijing}
  \country{China}}
\email{wangjingdong@baidu.com}

\author{Ziwei Liu}
\affiliation{%
  \institution{Nanyang Technological University}
  \city{Singapore}
  \country{Singapore}}
\email{ziwei.liu@ntu.edu.sg}

\author{Hideki Koike}
\affiliation{%
  \institution{Tokyo Institute of Technology}
  \city{Tokyo}
  \country{Japan}}
\email{koike@acm.org}

\renewcommand{\shortauthors}{Yasheng Sun, Hang Zhou, et al.}
\begin{abstract}

Previous studies have explored generating accurately lip-synced talking faces for arbitrary targets given audio conditions. However, most of them deform or generate the whole facial area, leading to non-realistic results. In this work, we delve into the formulation of altering only the mouth shapes of the target person. This requires masking a large percentage of the original image and seamlessly inpainting it with the aid of audio and reference frames. To this end, we propose the \textbf{Audio-Visual Context-Aware Transformer (AV-CAT)} framework, which produces accurate lip-sync with photo-realistic quality by predicting the masked mouth shapes. Our key insight is \emph{to exploit desired
contextual information provided in audio and visual modalities thoroughly with delicately designed Transformers}. Specifically, we propose a convolution-Transformer hybrid backbone and design an attention-based fusion strategy for filling the masked parts. It uniformly attends to the textural information on the unmasked regions and the reference frame. Then the semantic audio information is involved in enhancing the self-attention computation. Additionally, a refinement network with audio injection improves both image and lip-sync quality. Extensive experiments validate that our model can generate high-fidelity lip-synced results for arbitrary subjects.

\begin{figure}[t]
\begin{center}
\includegraphics[width=1.\linewidth]{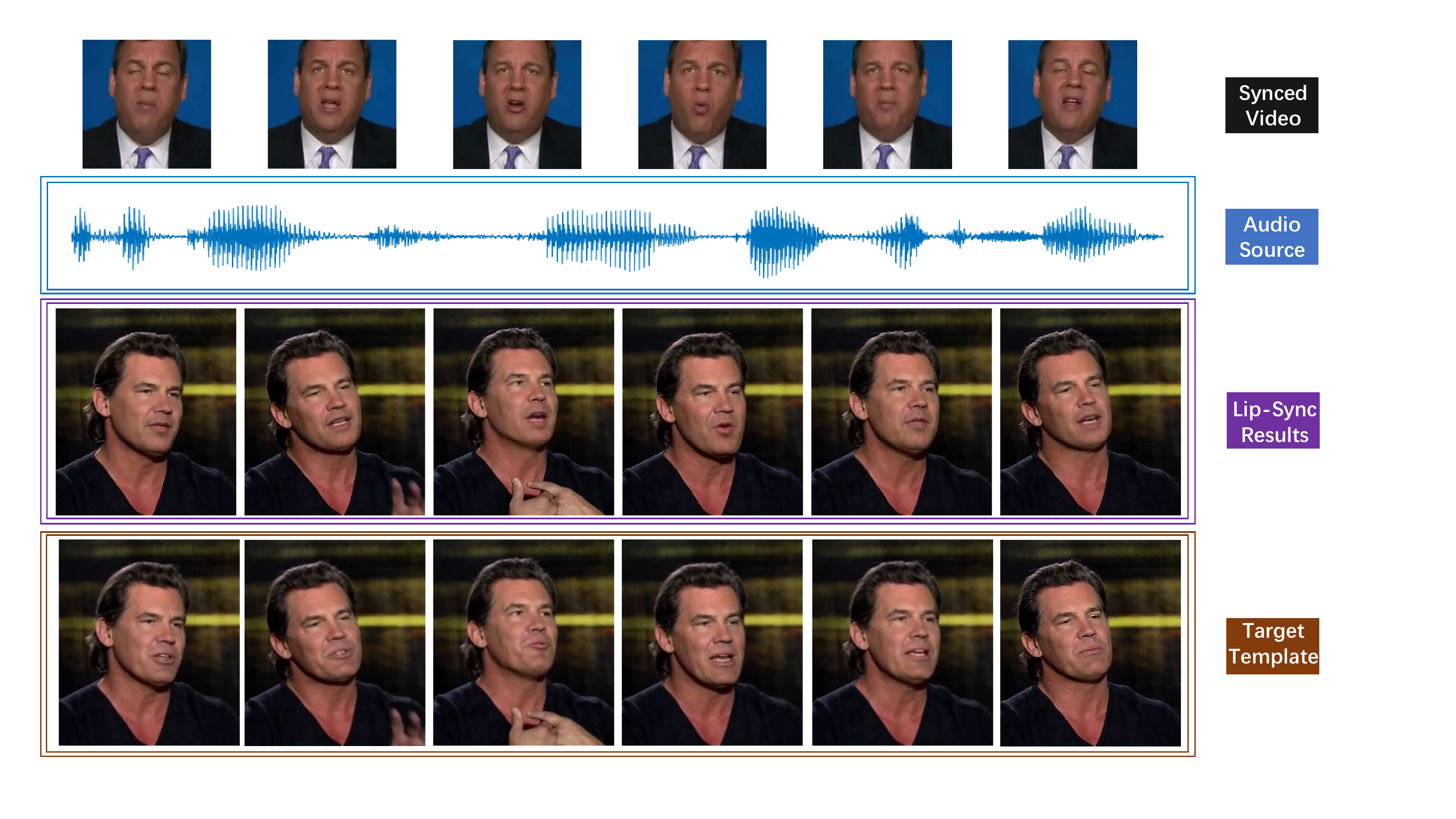}  
\end{center}
\caption{Qualitative results of our method. We focus on the task of high-fidelity person-agnostic lip-sync generation, which modifies the mouth shapes of any target template video according to the audio source. Here our lip-sync results should have the same mouth shape as the synced video to the audio source. The figures are selected from VoxCeleb~\cite{Nagrani17} and VoxCeleb2~\cite{Chung18b} \copyright Visual Geometry Group (CC BY).} 
\label{fig:block}
\end{figure}

\end{abstract}


\maketitle

\section{Introduction}
\label{sec:1}
Driving human mouth movements with speech audio is in great need in the field of virtual human creation, entertainment, the film industry, and the detection of deep fakes. The ability to drive arbitrary target video can vastly improve the applicability of such models. However, previous studies focusing on arbitrary targets mostly operate on the whole face~\cite{jamaludin2019you,zhou2019talking,zhou2020makelttalk,sun2021speech2talking,zhou2021pose,chen2018lip,chen2019hierarchical,vougioukas2019realistic,chen2020talking,song2018talking,wu2021imitating,wang2021audio2head,wang2022one,ji2022eamm}. They either 
reconstruct~\cite{zhou2019talking,zhou2021pose,zhou2020makelttalk} or warp~\cite{wang2021audio2head}] the whole head with generative models. As the generated moving head cannot be attached to a body, 
results under such a protocol can hardly be used in real-world applications.

Modifying only the mouth movements of a target video with audio input is an intriguing property that can be directly applied to scenarios like audio dubbing. {Previous researchers investigate the mouth modification problem mainly in case that relatively long videos are available for training~\cite{thies2020neural,song2020everybody,ji2021audio,suwajanakorn2017synthesizing}.} These methods suffer from two main drawbacks: 1) They cannot be directly applied to any-length videos, which limits their applicability. 2) They rely on intermediate structural priors such as 2D/3D landmarks or 3D models. The inaccuracy in such predictions would impede the audio-to-mouth prediction procedure.
Only a few studies~\cite{prajwal2020lip,park2022synctalkface} are proposed to tackle the problem of audio-driven lip-sync for an arbitrary subject. Wav2Lip~\cite{prajwal2020lip} specifically recovers the lower half of a face with speech and a reference frame. Though their mouth movements match nicely with the audio, they can only produce results with low quality, particularly around the mouth. 

We identify that such a setting of modifying an arbitrary target image with audio is a particular type of conditional masked image prediction~\cite{chang2022maskgit}, which we refer to as \emph{masked lip-sync prediction}. { Specifically, we mask most of the lower face. Two types of \emph{contextual} information, including the textural appearance information and the semantic mouth information, are needed for filling the masked areas. They can be extracted from three perspectives: \textbf{1)} The visible part of the masked image itself which contains pose and environment information; \textbf{2)} An appearance reference image; and \textbf{3)}  the semantic speech content information from audio. }
%
 Balancing such \emph{contextual information} requires modeling non-local intrinsic, non-aligned extrinsic, and cross-modal information. This is particularly challenging for convolutional networks that rely on deeper layers for fusing distant contexts, especially when the expected quality of the generated results is high. 
 { The objective of preserving more details from the reference images naturally contradicts the goal of mouth modifications with audio.}


In this work, we achieve masked lip-sync prediction for an arbitrary person through our  \textbf{Audio-Visual Context-Aware Transformer (AV-CAT)} and successfully produce high-quality results.
Our key insight is that  \emph{the desired semantic and appearance
contextual information provided in both audio-visual modalities can be thoroughly exploited with a delicately designed Transformer structure}.
We choose the backbone network inspired by recent advances on Vision Transformers~\cite{vaswani2017attention,dosovitskiy2020image,liu2021swin} and their applications in image modeling~\cite{bao2021beit,chen2022context,chang2022maskgit,esser2021taming}. As the attention mechanisms~\cite{wang2018non} can naturally treat information at arbitrary positions equally, Transformers have already been leveraged in generative tasks including super resolution~\cite{yang2020learning}, inpainting~\cite{zheng2021tfill,li2022mat,yu2021diverse} and image synthesis~\cite{esser2021imagebart,zhang2021styleswin,chang2022maskgit}. {They naturally suit our target of fusing the contextual information across frames and modalities. . 

In detail, we adopt a lightweight hybrid convolution-Transformer backbone structure to cope with the fusion challenge.} 
It takes advantage of both the inductive bias in convolution layers and explores patch-wise cross-frame and cross-modal information fusion in Transformers. 
We design a Cross-modal Contextual Fusion Transformer block (CCF-Transformer)  which consists of a cross-frame mutual attention and a cross-modal audio information fusion strategy based on basic Swin Transformer~\cite{liu2021swin} designs.  {While it accounts mostly for coarse-grained mouth movements, it fails to model delicate details.
To cope with this issue, we propose an audio-injected Refinement Network.} It is a UNet-like structure with modulated convolutions conditioned on audio. 
With this refinement module, both the generation quality and the mouth movement accuracy can be improved. Extensive experiments validate the effectiveness of each model. Through our delicately designed framework, our model is able to generate lip-synced results with high fidelity.

Our contributions can be summarized as follows: \textbf{1)} We leverage cross-frame and cross-modal context information for masked lip-sync prediction with delicately designed attention blocks. \textbf{2)} We propose the refinement network, which enhances the generated image quality and lip-sync results with the modulated injection of audio information. \textbf{3)} Extensive experiments validate that our Audio-Visual Context-Aware Transformer framework generates high-fidelity lip-sync results for arbitrary subjects.
\section{Related Work}

\subsection{Audio-Driven Lip-Synced Talking Face}
Driving a target portrait with speech audio has long been a popular research topic in the  computer graphics and computer vision  communities.
Generally, previous approaches can be classified into person-specific and person-agnostic settings.

\subsubsection{Person-Specific Lip-Sync Synthesis.} Most person-specific modeling methods generate high video quality~\cite{suwajanakorn2017synthesizing,thies2020neural,song2020everybody,ji2021audio,guo2021adnerf,lahiri2021lipsync3d,li2021write,liu2022semantic}. Some of them focus on altering the mouth areas for photo-realistic results~\cite{suwajanakorn2017synthesizing,thies2020neural,song2020everybody}. \citet{suwajanakorn2017synthesizing} design an audio-to-mouth module to synthesize high-quality talking face videos of Obama.  \citet{song2020everybody} introduce  3DMM~\cite{blanz1999morphable} to translate the speech content into the landmarks of the mouth area. \citet{thies2020neural} leverage a similar idea for mouth rendering.
On the other hand, \citet{guo2021adnerf} and \citet{liu2022semantic} use NeRF, \citet{ji2021audio} use 2d landmarks and \citet{lahiri2021lipsync3d} use neural rendering to synthesize the whole talking faces. All the above results enjoy high video quality, however, their methods cannot be applied to a person with few (seconds of) data, which limits their applicability.

\begin{figure*}[t]
\begin{center}
\includegraphics[width=0.97\linewidth]{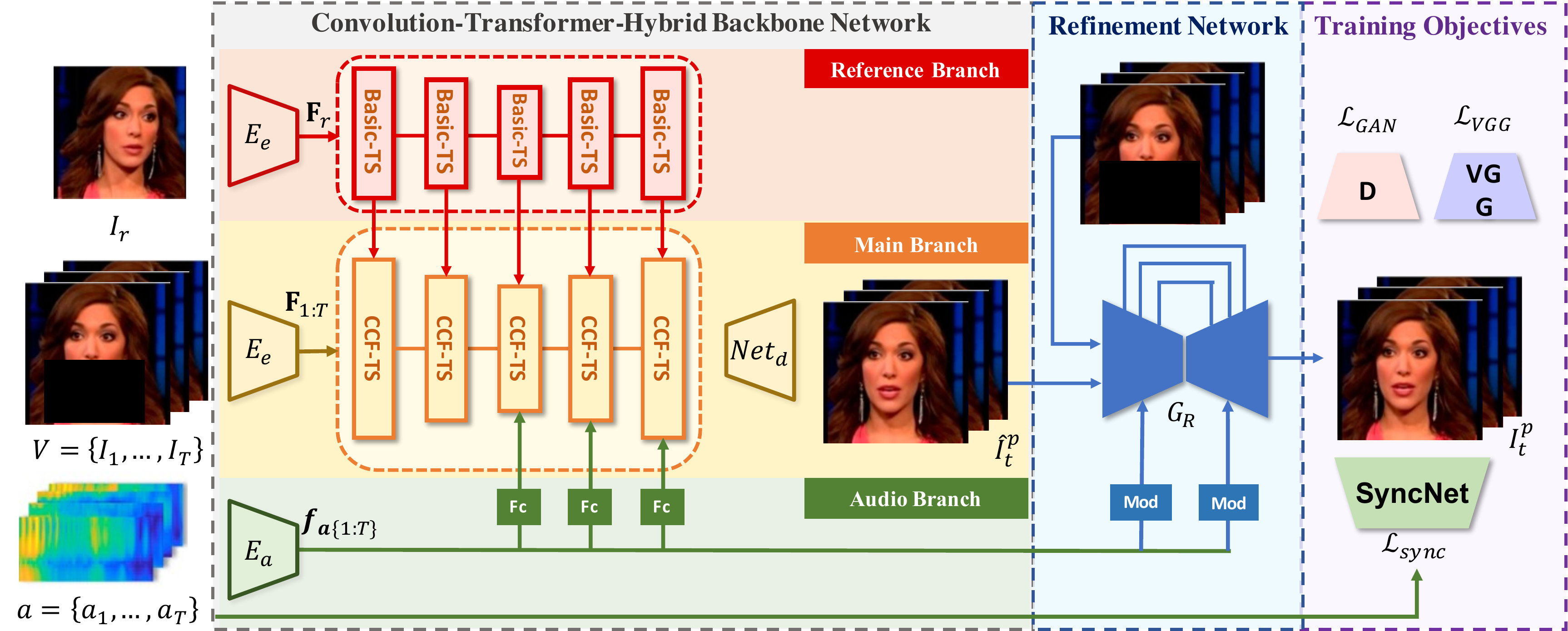}  
\end{center}
\caption{Overview of our \emph{Audio-Visual Context-Aware Transformer (AV-CAT)} framework. It consists of a convolution-Transformer-hybrid backbone network and a Refinement Network. The Reference image $I_r$ is processed in the reference branch with Basic Transformer stages (Basic-TS). The audio information is processed in the audio branch. Both this information is fused to the CCF-Transformer
stages (CCF-TS) in the main branch. The figures are selected from VoxCeleb~\cite{Nagrani17} \copyright Visual Geometry Group (CC BY).} 
\label{fig:pipeline}
\end{figure*}

\subsubsection{Person-Agnostic Lip-Sync Synthesis} 
Other methods focus on the person-agnostic setting where only one image or a short clip is provided as a reference. Speech2Vid~\cite{jamaludin2019you} studies the problem for the first time in an end-to-end manner. Then \citet{chen2019hierarchical} leverage 2D facial landmarks as guidance. \citet{zhou2020makelttalk} propose to use 3D landmark displacements and synthesize videos including head poses and blinks. \citet{zhou2021pose} choose to modularize audio-visual representations and manage to generate taking faces with different head poses. Particularly, \citet{yi2020audio} and \citet{chen2020talking} intend to generate more rhythmic realistic results with a short clip. However, methods modeling the whole head tend to change the background and body deformation, making them less realistic. 

Specifically, Wav2Lip~\cite{prajwal2020lip} focuses on the mouth areas and inpaints the lower half of the face using speech audio and another set of reference frames. {However, they rely on CNN-based UNet with skip-connections and concatenate audio information to the bottleneck of the network, which leads to blurry results.}
In the pursuit of high-quality and person-agnostic lip-sync, we adopt a similar setting as Wav2Lip and identify the problem as masked lip-sync prediction.
Our method can synthesize clearly more realistic results than previous person-agnostic methods. 

\subsection{Image Synthesis with Transformers}
The Transformer architecture~\cite{vaswani2017attention} has received growing interest from various tasks in computer vision~\cite{dosovitskiy2020image,bao2021beit,liu2021swin,esser2021taming,chang2022maskgit,li2022mat,he2021masked}. Recently, Transformer has also extended its reach to low-level vision tasks, such as image synthesis~\cite{van2017neural, esser2021taming} and image restoration~\cite{yang2020learning, yang2021implicit, zheng2021tfill,li2022mat,yu2021diverse}.


For conditional image restoration tasks (\textit{e.g., super-resolution and inpainting}), \citet{yang2020learning} propose a texture transformer network for image super-resolution by leveraging different high/low-resolution references. \citet{yang2021implicit} present a transformer-based framework for continuous image super-resolution with implicit function.
\citet{li2022mat} introduce a mask-aware transformer (MAT) for large-hole inpainting.
Similar to the Transformer-based approaches mentioned above, our AV-CAT also aims to fill in the masked region and synthesize high-fidelity images with reasonable textural appearance and correct semantic mouth movements. 

\section{Methodology}

In this section, we introduce our Audio-Visual Context-Aware Transformer (AV-CAT) as illustrated in Fig.~\ref{fig:pipeline}. {The whole framework contains a  convolution-Transformer-hybrid backbone network that accounts for information fusion at coarse level and a convolutional Refinement Network that fixes details. }
We will first introduce our masked lip-sync prediction training setting in Section~\ref{sec:3.1}. Then we detailedly depict the design of the backbone network and information fusion strategies. Finally, we introduce the Refinement Network and the training objectives.

\subsection{Training Setting}
\label{sec:3.1}

Our masked lip-sync prediction setting is similar to Wav2Lip~\cite{prajwal2020lip}. Given a training video $\textbf{V} = \{I_1, \dots, I_T\}$ with corresponding audio clip $\textbf{a} = \{a_1, \dots, a_T\}$, {our model processes one target frame $I_t \in \textbf{V}$ at a time. We mask out most areas on the lower half of  $I_t$ to $I^m_t = M * I_t$ with a fixed mask $M$.} The training goal is to recover $I_t$ with the corresponding audio $a_t$. To recover the mouth-shape irrelevant information such as textures on the face  as well as the masked hair and backgrounds,  a reference frame $I_r \in \textbf{V}$ is involved. We expect the reference frame to provide textual information only and avoid exacting the mouth shapes directly from it. Thus it is sampled from the same video but at a different timestamp. The reference frame and the target can be selected as the same during testing.

\subsection{Audio-Visual Context-Aware Transformer} 
The convolution-Transformer-hybrid backbone network consists of three branches: the \emph{main branch}, \emph{reference branch} and the \emph{audio branch}. They target processing the target frame, reference frame, and audio information, respectively. The main branch consists of a convolutional encoder head, 5 \emph{Cross-Modal Contextual Fusion Transformer (CCF-Transformer)} stages, and a convolutional decoder. Specifically, the CCF-Transformer is built upon the \emph{Basic Transformer} blocks modified from Swin Transformers~\cite{liu2021swin}.
We design our reference branch without involving extra parameters. It uses the same convolutional encoder $E_e$ for feature map encoding. Moreover, it consists of 5 stages of Basic-Transformer blocks which share all learnable weights with CCF-Transformer's sub-modules. The audio branch encodes an audio feature $f_a$ with encoder $E_a$. Both $f_a$ and intermediate features of the Basic-Transformer blocks are sent into the CCF-Transformer blocks for information fusion.

\subsubsection{Convolutional Encoder and Decoder.} We choose to preserve 4 layers of convolution operations for the low-level representation downsampling in encoder $E_e$ and another 4 for upsampling in decoder $Net_d$. It has been verified that the inductive bias at early layers is effective for Transformer learning~\cite{raghu2021vision,xiao2021early,li2022mat}. Specifically, the convolutional encoder $E_e$ processes an image to a feature map $\textbf{F}$ with $1/4$ the size of $I_t$. 
It encodes the target and the reference frame to feature maps $\textbf{F}_t$ and $\textbf{F}_r$ respectively.

\subsubsection{Basic Transformer Structure.} The reference feature map $\textbf{F}_r \in \mathbb{R}^{h\times w \times C_0}$ is then sent into our Basic Transformer of 5 stages, each containing several blocks. It is built based on Swin Transformer~\cite{liu2021swin} with modifications on the hierarchical representations inspired by~\cite{li2022mat}, including the removal of the layer normalization. While Swin Transformer downsamples the feature patches at each stage by fusing neighboring $2 \times 2$ patches, we consider a downsample-upsample strategy given our generative task property. At the beginning of the first three stages, we downsample the feature map with a stride-2 convolution with $2 \times 2$ kernel size. Differently, we upsample the feature maps at the last two stages with transposed convolutions.

The self-attentions is computed within local windows of size $w_w \times w_w$ as performed in~\cite{liu2021swin}, which saves computational cost. We adopt the standard multi-head self-attention (MSA) in the Basic Transformer blocks:
\begin{align}
    \text{Attention}(\textbf{Q}, \textbf{K}, \textbf{V}) = \text{Softmax}({\textbf{Q}\textbf{K}^{\text{T}}}/{\sqrt{d_k}})\textbf{V},
\end{align}
where \textbf{Q}, \textbf{K}, \textbf{V} are all tokens within the same window.
After each attention operation, the window is shifted by $(\lfloor w_w/2 \rfloor, \lfloor w_w/2 \rfloor)$ pixels.

We visualize the architecture of one Basic Transformer block in Fig.~\ref{fig:block} (a). Note that all the learnable parameters in the Basic Transformer are jointly learned with CCF-Transformer. This is to ensure the same semantics among both Transformer blocks and memory saving. The block is modified according to~\cite{li2022mat}.

\subsubsection{Cross-Modal Contextual Fusion Transformer Blocks.} The target's feature map $\textbf{F}_t \in \mathbb{R}^{h\times w \times C_0}$ is sent into CCF-Transformer which shares similar setups as the Basic Transformer, e.g., the resolution of each stage, the sliding window operation, and the computation of self-attentions. Differently, the CCF-Transformer receives extra contextual information from the reference branch and the audio branch: 1) The cross-frame reference information provided from the Basic Transformer blocks. It is fused to CCF-Transformer blocks. 2) The cross-modal audio information is mapped to the last 3 stages through a fully connected layer.

Detailedly, at each block of the CCF-Transformer, an additional multi-head mutual attention (MMA) is developed in parallel with MSA between the target feature map $F_{t\{k, l\}}$ and reference feature map $F_{r\{k, l\}}$, where $l$ and $k$ denote the $l$th block of the $k$th stage. In the MMA, the tokens from $F_{t\{k, l\}}$ are treated as queries and $F_{r\{k, l\}}$ serve as keys and values. This is to retrieve desired appearance information for inpainting missing areas. Then, the MMA and MSA results are concatenated together and mapped to the original feature dimension by a fully connected layer.

Then for the last three stages of CCF-Transformer, the audio feature $f_a$ is mapped to $f_{a\{k, l\}}$ and sent into the Transformer blocks to operate on the fused feature mentioned above.
The overall computation of one CCF-Transformer block can be written as:
\begin{align}
    &z^i_{\{k, l\}} = \text{FC}_1(\text{Cat}[\textbf{MSA}(F_{t\{k, l\}}),  \textbf{MMA}(F_{r\{k, l\}}, F_{t\{k, l\}})]), \\
      &z^{i}_{\{k, l\}} = \text{FC}_2(f_{a\{k, l\}} + z^{i}_{\{k, l\}})~~(\text{if k > 2}), \\
    &z^{ii}_{\{k, l\}} = z^{i}_{\{k, l\}} + F_{t\{k, l\}}, \\
    &F_{t\{k, l+1\}} = \text{MLP}(z^{ii}_{\{l,k\}}) + z^{ii}_{\{l,k\}},
\end{align}
which renders the feature map $F_{t\{k, l+1\}}$ of the next block. In the equations, $\text{Cat[]}$ denotes feature concatenation and $\text{FC}_i$ are different fully connected layers that are not shared with Basic Transformer blocks. 
Notably, the intuition for only injecting three layers of audio information is that audio usually interacts with high-level visual features~\cite{zhou2019talking,prajwal2020lip}. Experiments show that injecting audio into all 
stages makes little difference.

The final output of the CCF-Transformer is sent into the convolutional decoder $Net_d$ to predict an intermediate result $\hat{I}^p_t$.

\begin{figure}[t]
\begin{center}
\includegraphics[width=0.92\linewidth]{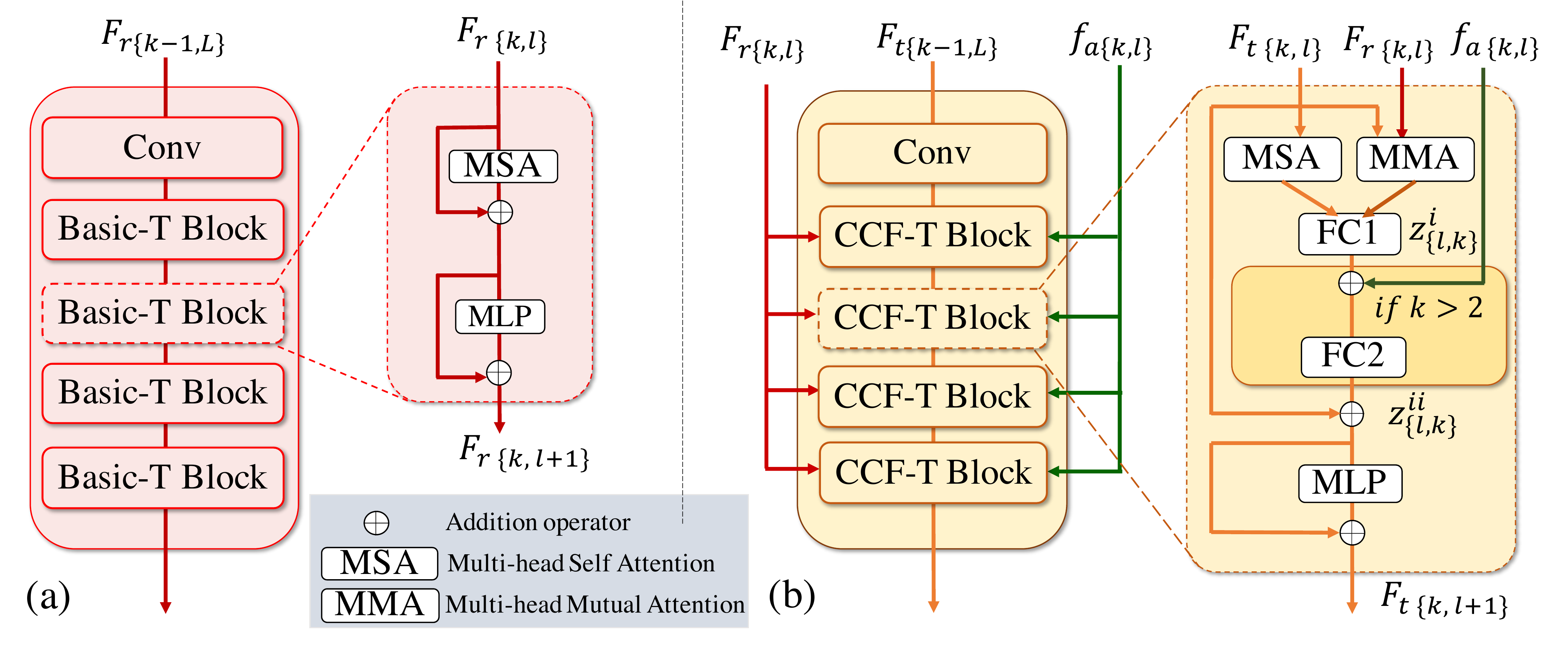}  
\end{center}
\caption{The architecture of our modified transformer blocks. (a) shows one block of the Basic Transformer in one stage, and (b) shows one block of the CCF-Transformer.} 
\label{fig:block}
\end{figure}


\subsection{Refinement Network}
{Note that though the convolution-Transformer-hybrid backbone (CTH backbone) manages to fuse effective information, it fails at handling fine-grained details. On the other hand, we identify that UNet-like~\cite{ronneberger2015u} structures are suitable for improving roughly good results. 
Thus, we propose the
Refinement Network $G_R$.} 
It takes the $\hat{I}^p_t$ predicted from the CTH backbone and the original masked target $I^M_t=M * I_t$ for generating more harmonic results with higher quality. 
Specifically, the input of the Refinement Network is $\hat{I}^{p'}_t = M * I_t + (\textbf{1} - M) * \hat{I}^p_t$ and it predicts $I^p_t = G_R(\hat{I}^{p'}_t)$.

Specifically, motivated by previous research on talking head generation that audio features can be effectively expressed with modulated convolution~\cite{zhou2021pose}, we adopt a similar protocol {for enhancing cross-modal learning}. We map audio feature $f_a$ to a style vector $s$ for each  convolution operation in the UNet.
For each value $W_{mnq}$ in any convolution kernel weight $W$, where $m$ channel-wise position of the input, $n$ is its position on the  output channel and $q$ stands for spatial location, we modulate it according to $s$'s given its channel-wise position $m$: 
%
\begin{align}
    W'_{mnq} = \frac{s_m \cdot W_{mnq}}{\sqrt{{\sum}_{m,q} (s_m \cdot W_{mnq})^2 + \epsilon}},
\end{align}
where $\epsilon$ is set as a small value to avoid numerical errors.

\subsection{Learning Objectives}
The Learning objectives are mainly the VGG and GAN losses that are leveraged in various generation tasks~\cite{pix2pix2017,wang2018pix2pixHD,zhou2021pose}. Normally the VGG loss is used for  perceptual similarity. Here we extend the VGG loss by computing from the first layer of the convolution to the $N_{vgg}$'s layer. This accounts for low-level information reconstruction. The loss functions are written as:
\begin{align}
\label{eq:4}
\begin{split}
    \mathcal{L}_{\text{GAN}} =& ~ \underset{\text{G}}{\text{min}}\underset{{D}}{\text{max}}(\mathbb{E}_{I_{t}}[\log {D}(I_{t})]  \\ 
    & + \mathbb{E}_{I^p_t}[\log (1 - {D}(I^p_t))]), \\
\end{split}\\
\label{eq:6}
    & \mathcal{L}_{\text{vgg}} = \sum_{n=1}^{N_{P}}{\|\text{VGG}_n({I}_{t}) - \text{VGG}_{n}(I^p_t) \|_1 },
\end{align}
where $D$ denotes a discriminator.

Besides, we also re-train the SyncNet~\cite{chung2016out} discriminator for constraining temporal consistency and audio-lip synchronization as performed in Wav2Lip~\cite{prajwal2020lip} .
It takes 5 consecutive predicted frames $I^p_{t:t+5}$ as input. The loss is carried out between $I^p_{t:t+5}$ and their corresponding audio representations $a_{t: t+5}$.
\begin{align}
    \mathcal{L}_{sync} = \text{SyncNet}(åI^p_{t:t+5}, a_{t: t+5}).
\end{align}
The overall learning objective can be summarized as:
\begin{align}
    \mathcal{L}_{all} = \mathcal{L}_{\text{GAN}} + \lambda_{vgg} \mathcal{L}_{VGG} + \lambda_{sync}\mathcal{L}_{sync},
\end{align}

\begin{figure*}[t]
\begin{center}
\includegraphics[width=0.93\linewidth]{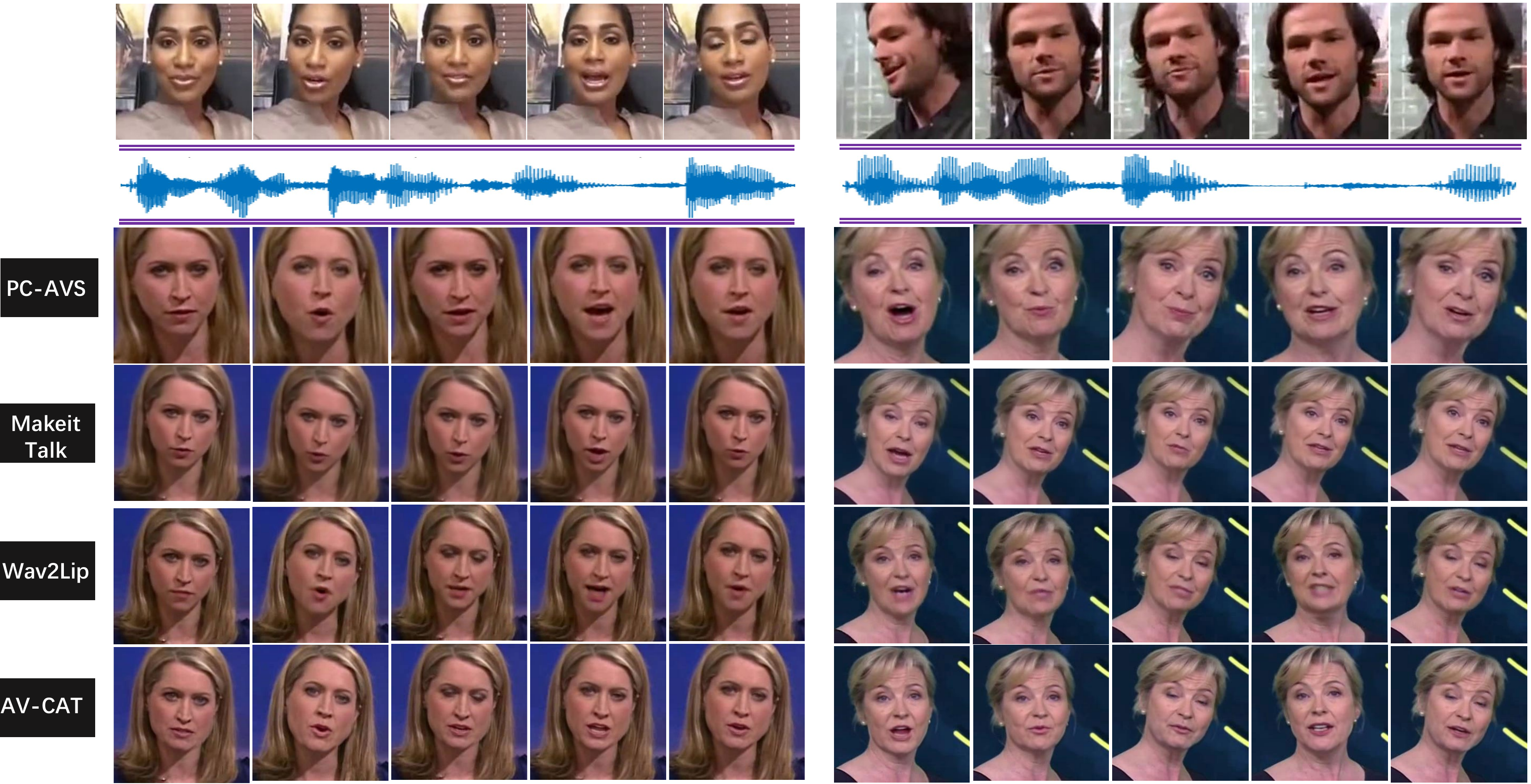}  
\end{center}
\caption{\textbf{Qualitative Results}. The top row shows the corresponding videos of the input audio. PC-AVS~\cite{zhou2021pose}  produces accurate lips, it is constrained to the cropped area. MakeitTalk~\cite{zhou2020makelttalk} fails to generate accurate mouth shape and lacks head dynamics. Wav2Lip~\cite{prajwal2020lip} syncs well with audios, however, the mouth areas are blurry with their method. Our AV-CAT generates realistic results with synced mouth movements. The figures are selected from VoxCeleb \copyright Visual Geometry Group (CC BY) and LRW \copyright BBC.} 
\label{quali}
\end{figure*}

\setlength{\tabcolsep}{6pt}
\begin{table*}[t] 
\begin{center}  \caption{\textbf{Quantitative results on LRW and VoxCeleb.} For LMD the lower the better, and the higher the better for other metrics.}

\label{table:exp1}
\begin{tabular}{lcccccccccc}
\toprule
 & \multicolumn{4}{c}{LRW~\cite{chung2016lip}}& \multicolumn{4}{c}{VoxCeleb~\cite{Nagrani17}} \\
\cmidrule(lr){2-5} \cmidrule(lr){6-9}
Method & $\text{SSIM}  \uparrow $& $\text{PSNR}  \uparrow $& LMD $\downarrow $  & $\text{Sync}_{conf}  \uparrow $ & $\text{SSIM}  \uparrow$& $\text{PSNR}  \uparrow $ & LMD $ \downarrow$ & $\text{Sync}_{conf}  \uparrow $\\

\midrule  
Wav2Lip~\cite{prajwal2020lip} &0.937 &34.25 & \textbf{2.54} &\textbf{6.7} &0.885 & 32.34 & \textbf{8.44} & 5.2 \\
MakeitTalk~\cite{zhou2020makelttalk}  &0.690 &31.09 & 5.03 &3.1 &0.814 & 29.51 &29.13 &2.3 \\
PC-AVS~\cite{zhou2021pose} &0.895 &33.87  & 3.04 & 6.2 &0.865 & 32.67 &8.91 &\textbf{5.4} \\
Ground Truth & 1.000 &100.00 & 0.00 & 6.5 & 1.000 & 100.0 & 0.00 &5.4 \\
\hline
\textbf{AV-CAT (Ours)}  & \textbf{0.938} & \textbf{36.34} & 2.83 &6.0 &\textbf{0.889} & \textbf{33.41} & 8.64 & 5.1 \\
\bottomrule
\end{tabular}
\end{center}

\end{table*}

\section{Experiments}
\paragraph{Datasets.} All the experiments are conducted on two audio-visual datasets LRW~\cite{chung2016lip} and VoxCeleb~\cite{Nagrani17}. Both of them are composed of in-the-wild data.
\begin{itemize}
    \item \textbf{LRW~\cite{chung2016lip}.} LRW is an audio-visual dataset collected from BBC news with high visual quality. This dataset consists of 1,000 utterances in total covering 500 words, and each utterance lasts about 1 second.
    \item \textbf{VoxCeleb~\cite{Nagrani17}.} VoxCeleb is a large-scale audio-visual dataset including 1,251 celebrities. The videos in VoxCeleb are all longer than 3 seconds but distributed in different visual qualities. 
\end{itemize}
In our experiments, we follow their train/test split strategies to train and test our AV-CAT framework. Our method is trained on LRW and about one-fifth of VoxCeleb.

\paragraph{Implementation Details.}
We process all videos at 25 fps at the size of $256 \times 256$. {We adopt a similar setting as \cite{prajwal2020lip} but leave 16 pixels more on the left and right sides unmasked. Thus 43.75\% of the input image is masked for every frame. All faces in all frames are aligned according to the eyes during training and inference. } We then follow the audio processing  of~\cite{zhou2021pose}, with 16kHz sampling rate and converted to mel-spectrograms with $\text{FFT}$ window size 1280, hop length 160, and 80 Mel filter-banks. The mel-spectrogram corresponds to the time-length of 5 frames with the target frame in the middle is sampled as the condition. At the 5 stages in both Basic Transformer and CCF-Transformer, there are \{2, 3, 4, 3, 2\} blocks. The window
sizes are
\{8, 16, 16, 16, 8\}, respectively. The $\lambda_{vgg}$ is  set as 1 and $\lambda_{sync}$ empirically set as 0.0001. {Note that the network processes each frame independently during inference. The consecutive frames are only used in SyncNet during training.}

\paragraph{Comparing Methods.}
We compare our AV-CAT framework with three  person-agnostic state-of-the-art methods, MakeitTalk~\cite{zhou2020makelttalk}, PC-AVS~\cite{zhou2021pose} and Wav2Lip~\cite{prajwal2020lip}.  
MakeitTalk and PC-AVS all synthesize talking heads with head pose movements. Wav2Lip~\cite{prajwal2020lip} shares the same setting as ours.

\subsection{Quantitative Evaluation} 
The evaluation is constructed in a self-reconstruction manner.
Given that it is not fair to use the whole ground truth information directly as a reference input, we uniformly select the first frame as the reference frame for all methods to ensure that it cannot provide lip-sync information. 
\subsubsection{Evaluation Metrics.}
We evaluate our AV-CAT on both generation quality and lip synchronization.
For the generation quality, we follow previous studies~\cite{zhou2019talking,chen2019hierarchical,zhou2021pose} and adopt the commonly used SSIM~\cite{wang2004image}, PSNR as metrics. 
In terms of lip synchronization, we follow \cite{chen2019hierarchical,zhou2021pose} to use the landmarks distances around the mouth region (LMD) to account for the generated mouth shape. Then we employ the confidence score of SyncNet~\cite{chung2016out} to evaluate the synchronization quality as performed in~\cite{vougioukas2019realistic,zhou2021pose}. Note that we use the officially SyncNet as a metric instead of our self-trained for the evaluation.

\begin{figure}[t]
\begin{center}
\includegraphics[width=0.9\linewidth]{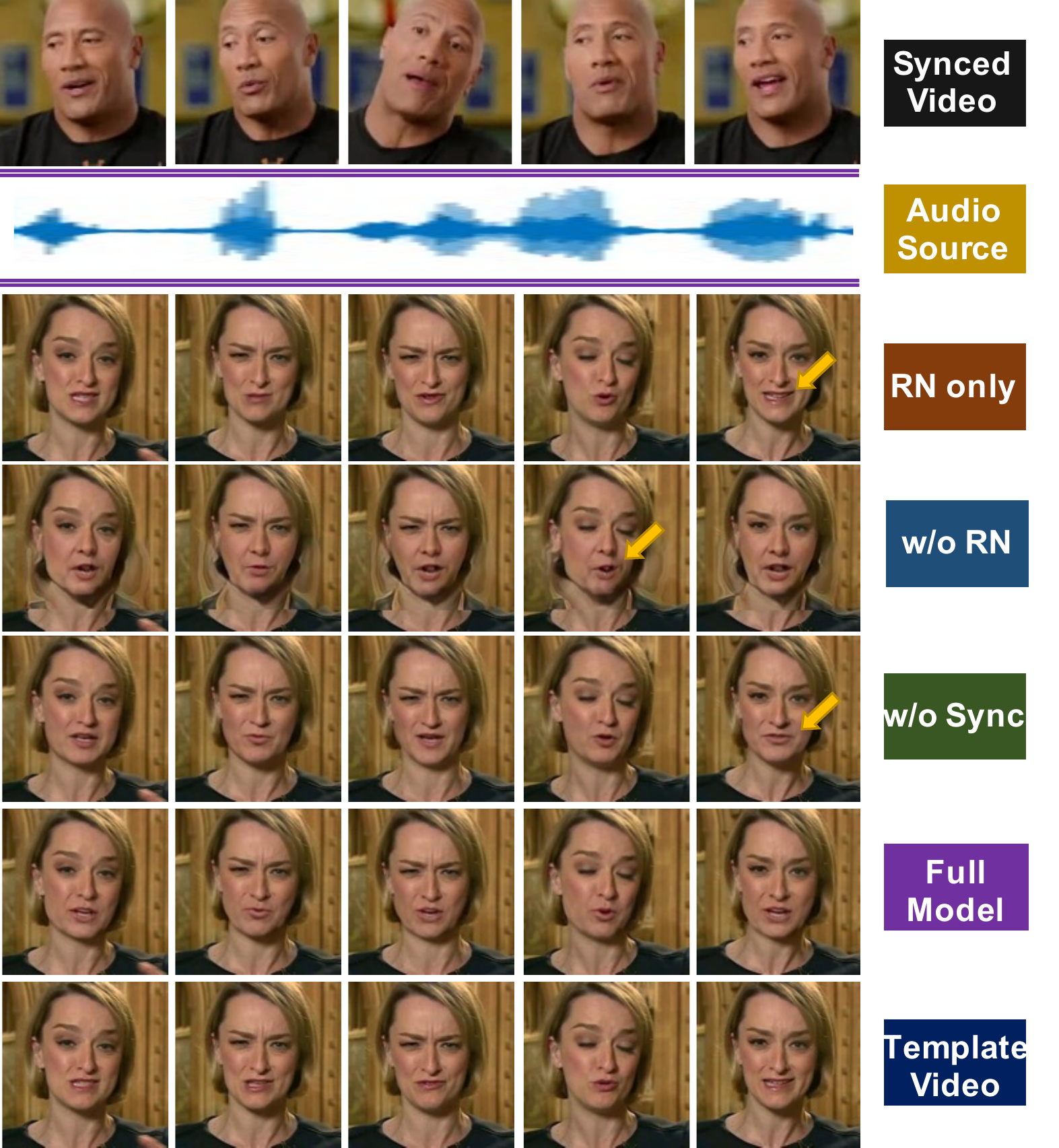}  
\end{center}
\caption{\textbf{Ablation studies with visual results}. 
Only the full model can synthesize videos with both correct mouth movement and face shape. The figures are selected from VoxCeleb \copyright Visual Geometry Group (CC BY) and LRW \copyright BBC.} 
\label{ablation}
\end{figure}

\subsubsection{Evaluation Results.}
We perform quantitative comparisons on the test set of LRW~\cite{chung2016lip} and VoxCeleb~\cite{Nagrani17} datasets. The results are summarized in Table~\ref{table:exp1}. 
Compared to existing methods, our AV-CAT achieves significant improvement in the generation quality. 
Meanwhile, our AV-CAT also achieves comparable performance with the state-of-the-art lip-sync performance on both datasets.
Our LMD score is slightly worse than Wav2Lip, possibly caused by the observation that we tend to generate more exaggerated lip movements.
This stands the same for the SyncNet scores. It is possible that our self-trained SyncNet discriminator does not perform as well as the one used in Wav2Lip. 

{Particularly, we argue there is generally a trade-off between perceptual quality and synchronization and our model balances it better than traditional CNN-based models. We also re-train a Wav2Lip model with the same setting and dataset as ours (denoted as Wav2Lip-L) and find it achieves worse performance than the original model. Both the limited receptive fields and the skip-connections at higher resolutions restrict the network’s ability.}
%

\subsection{Qualitative Evaluation}
We also conduct a qualitative comparison, including the results in Fig.~\ref{quali} and a subjective evaluation. 
For video samples, please refer to our supplementary materials\footnote{Demo videos are available at \url{https://hangz-nju-cuhk.github.io/projects/AV-CAT.}
}. 
The first line of Fig~\ref{quali} shows the frames of the original videos that correspond to input audio. 
As shown in Fig~\ref{quali},
MakeitTalk fails to predict accurate mouth shapes due to the prediction error from the 3D landmarks. Moreover, its generated head movements are relatively subtle compared with original frames.
On the other hand, PC-AVS synthesizes basically correct mouth shapes, but their results are limited in the cropped facial areas, making them non-realistic.
%
Wav2Lip synthesizes satisfactory lip-sync performance. However, the inpainted mouth region is always blurry, given that they can only handle images of low qualities.
Our AV-CAT generates high fidelity results with much sharper mouth boundaries and clearer textures.


\subsubsection{User Study.}
We additionally perform a user study, where 15 participants are invited to evaluate results generated by our AV-CAT and three comparing methods.
20 videos are sampled from VoxCeleb and 20 from LRW. We randomly select the driving audios from the test set under the cross-video driving setting.
Following the commonly used Mean Opinion Scores (MOS) protocol, the participants are required to rate (1-5) considering the following factors: (1) \textbf{Lip-sync quality}; (2) the \textbf{perceptual quality} and (3) the \textbf{realness} of generated videos.
The results are reported in Table~\ref{table:MOS}. It can be seen that our results reach the best on all three perspectives, which proves the high quality of our method.

\subsection{Ablation Studies}
We conduct ablation studies further to demonstrate the contributions of different components in our AV-CAT. 
Specifically, we construct three variants by removing the specified components: 
\textbf{1)} First, to evaluate the effectiveness of the transformer architecture, we remove the CTH backbone, and directly feed the masked frames and references into the Refinement Network (denoted as ``{RN only}''). Since our Refinement Network is designed based on U-Net and audio injection. 
{We also try to replace the CTH backbone network to the larger version of Wav2Lip in our model (Wav2Lip-L + RN).}
\textbf{2)} Then in order to evaluate the contribution of the Refinement Network, we remove the Refinement Network, and evaluate the intermediate output of the transformer backbone (denoted as ``w/o RN''). 
\textbf{3)} Finally, to evaluate the contribution that SyncNet makes to the lip-synchronization quality, we train another model without using the $\mathcal{L}_{sync}$ (denoted as ``w/o Sync'').

The numerical results are shown in Table~\ref{table:ablation} and the visual results are shown in Fig.~\ref{ablation}, respectively. 
Compared to our Full Model, 
``RN only'' achieves the worst results on LMD and SyncNet scores due to the redundant mouth shape information from the reference frame (the results are close to the target template as shown in Fig.~\ref{ablation}). 
``W/o RN'' obtains degraded performance in all metrics, which represents the inconsistent texture of the mouth region (see second row of Fig.~\ref{ablation}).
{The effectiveness of our CTH backbone network can further be shown as the results combining Wav2Lip-L with RN perform worse than our full model.}
``w/o Sync'' fails to generate synchronized lip movements. 
Both the numerical results and visual results demonstrate the effectiveness of our components in  AV-CAT.

\setlength{\tabcolsep}{3pt}
\begin{table}[t] 
\begin{center}

\caption{\textbf{User study measured by Mean Opinion Scores.} Larger is higher, with the maximum value to be 5.}

\label{table:MOS}
\begin{tabular}{ccccc}

\hline

MOS on $\setminus$ Method & PC-AVS & Wav2Lip & MakeitTalk & \textbf{AV-CAT} \\
\noalign{\smallskip}
\hline

Lip-Synch Quality & 3.54 & 3.68 & 2.52 & \textbf{3.82} \\
Generation Quality& 3.94 & 3.85 & 2.95 & \textbf{4.30}\\
Video Realness    & 3.56 & 3.48 & 3.37 & \textbf{4.18}\\
\hline
\end{tabular}
\end{center}

\end{table}

\setlength{\tabcolsep}{8pt}
\begin{table}[t] \footnotesize

\caption{\textbf{Ablation study with quantitative comparisons on VoxCeleb.} The results are shown when we vary the accessibility of the Transformer network, refinement network, and the optimization objective.}
\label{table:ablation}

\begin{center}  
\begin{tabular}{lcccccc} 
\toprule
Method & $\text{SSIM}  \uparrow$ & \text{PSNR} $\uparrow $  & LMD $\downarrow $ & $\text{Sync}_{conf}$$\uparrow $\\

\midrule  
RN only &0.812 & 30.61 & 10.68 & 3.9 \\
w/o RN &0.748 & 29.31 & 10.16 & 4.9  \\ 
Wav2Lip-L + RN & 0.874 & 32.29 & 8.72 & 4.8 \\
w/o Sync &0.868 & 32.93 & 9.34 & 4.7 \\

\textbf{Full Model} &\textbf{0.889} & \textbf{33.41} & \textbf{8.64} &\textbf{5.1} \\
\bottomrule
\end{tabular}
\end{center}

\end{table}

\section{Conclusion}

In this work, we present the Audio-Visual Context-Aware Transformer (AV-CAT), which produces high-quality lip-synced talking faces. We highlight several  important proprieties. 1) We  formulate the problem in a masked lip-sync prediction manner, and propose to learn cross-frame and cross-modal context information with Transformers. 2) Our design shows that audios can modulate UNet structures for quality improvement. 3) Our results are validated to be more realistic than the previous state of the arts.

{
\paragraph{Limitations and Future Work.} \textbf{1)} 
The model is not sensitive to certain consonants. Possibly, the data are not evenly distributed. 
The state of the reference image could also slightly affect the results. \textbf{2)} Moreover, our model cannot handle delicate details such as mimicking the speaking style of a specific person or casting shadows from the jaw. \textbf{3)} More advanced audio representations~\cite{baevski2020wav2vec,chen2022wavlm} can be involved in future studies.

\paragraph{Ethical Considerations.} Our method might be leveraged for malicious uses such as creating deepfakes. We will restrict the license of our model to research use only and share it with the deepfake detection community. We will also try actively adding watermarks into the generation process for deepfake identification.}  
\bibliographystyle{ACM-Reference-Format}
\bibliography{sample-base}

\appendix

\end{document}